\documentclass[a4paper,fleqn]{cas-dc}

\usepackage[authoryear,longnamesfirst]{natbib}

\usepackage{url}

\usepackage{float}
\usepackage{placeins}
\usepackage{multicol}
\def\tsc#1{\csdef{#1}{\textsc{\lowercase{#1}}\xspace}}
\tsc{WGM}
\tsc{QE}
\begin{document}
\let\WriteBookmarks\relax
\def\floatpagepagefraction{1}
\def\textpagefraction{.001}

\shorttitle{Leveraging LID for Code-Mixed Data}    

\shortauthors{G. Takawane, A. Phaltankar, V. Patwardhan, A. Patil et al}  

\title [mode = title]{Leveraging Language Identification to Enhance Code-Mixed Text Classification}  

% \tnotemark[<tnote number>] 



 \author[1,2]{Gauri Takawane}
% % [<options>]

% % Corresponding author indication
 \cormark[1]

% % Footnote of the first author
% % \fnmark[<footnote mark no>]

% % Email id of the first author
 \ead{gauri.takawane@gmail.com}

% % URL of the first author
% % \ead[url]{<URL>}

% % Credit authorship
% % eg: \credit{Conceptualization of this study, Methodology, Software}
% % \credit{<Credit authorship details>}

% % Address/affiliation
 \affiliation[1]{,
             Department={Department of Computer Engineering,    },
             organization={SCTR'S Pune Institute of Computer Technology},
             city={Pune},
% %          citysep={}, % Uncomment if no comma needed between city and postcode
             country={India}}

 \affiliation[2]{,
          %   Department={Department of Computer Engineering,    },
             organization={L3Cube Pune},
             city={Pune},
% %          citysep={}, % Uncomment if no comma needed between city and postcode
             country={India}}

\author[1,2]{Abhishek Phaltankar}
% % [<options>]

% % Corresponding author indication
 \cormark[1]

% % Footnote of the first author
% % \fnmark[<footnote mark no>]1

% % Email id of the first author
 \ead{avp1510@gmail.com}

% % URL of the first author
% % \ead[url]{<URL>}

% % Credit authorship
% % eg: \credit{Conceptualization of this study, Methodology, Software}
% % \credit{<Credit authorship details>}

\cortext[cor1]{Equal Contribution}

% %--------------------------------------------------------------

 \author[1,2]{Varad Patwardhan}
% % [<options>]

% % Corresponding author indication
 \cormark[1]

% % Footnote of the first author
% % \fnmark[<footnote mark no>]

% % Email id of the first author
\ead{varadp2000@gmail.com}

% % URL of the first author
% % \ead[url]{<URL>}

% % Credit authorship
% % eg: \credit{Conceptualization of this study, Methodology, Software}
% % \credit{<Credit authorship details>}

% %--------------------------------------------------------------

 \author[1,2]{Aryan Patil}
% % [<options>]

% % Corresponding author indication
 \cormark[1]

% % Footnote of the first author
% % \fnmark[<footnote mark no>]

% % Email id of the first author
\ead{aryanpatil1503@gmail.com}

% % URL of the first author
% % \ead[url]{<URL>}

% % Credit authorship
% % eg: \credit{Conceptualization of this study, Methodology, Software}
% % \credit{<Credit authorship details>}

% %--------------------------------------------------------------

 \author[2,3]{Raviraj Joshi}
% % [<options>]

% % Corresponding author indication
 \cormark[1]

% % Footnote of the first author
% % \fnmark[<footnote mark no>]

% % Email id of the first author
 \ead{ravirajoshi@gmail.com}

% % URL of the first author
% % \ead[url]{<URL>}

% % Credit authorship
% % eg: \credit{Conceptualization of this study, Methodology, Software}
% % \credit{<Credit authorship details>}

% % Address/affiliation
 \affiliation[3]{,
             Department={Department of Computer Science and Engineering,},
             organization={Indian Institute of Technology Madaras},
             city={Chennai},
% %          citysep={}, % Uncomment if no comma needed between city and postcode
             country={India}}

% %--------------------------------------------------------------

 \author[1]{Mukta S. Takalikar}
% % [<options>]

% % Corresponding author indication
% % \cormark[<corr mark no>]

% % Footnote of the first author
% % \fnmark[<footnote mark no>]

% % Email id of the first author
 \ead{mstakalikar@pict.edu}

% % URL of the first author
% % \ead[url]{<URL>}

% % Credit authorship
% % eg: \credit{Conceptualization of this study, Methodology, Software}
% % \credit{<Credit authorship details>}

% % Corresponding author text
% % \cortext[1]{Corresponding author}

% % Footnote text
% % \fntext[1]{}

% % For a title note without a number/mark
% %\nonumnote{}

\begin{abstract}
The usage of more than one language in the same text is referred to as Code Mixed. It is evident that there is a growing degree of adaption of the use of code-mixed data, especially English with a regional language, on social media platforms. Existing deep-learning models do not take advantage of the implicit language information in the code-mixed text. Our study aims to improve BERT-based models’ performance on low-resource Code-Mixed Hindi-English Datasets by experimenting with language augmentation approaches. We propose a pipeline to improve code-mixed systems that comprise data preprocessing, word-level language identification, language augmentation, and model training on downstream tasks like sentiment analysis. For language augmentation in BERT models, we explore word-level interleaving  and post-sentence placement of language information. We have examined the performance of vanilla BERT-based models and their code-mixed HingBERT counterparts on respective benchmark datasets, comparing their results with and without using word-level language information. The models were evaluated using metrics such as accuracy, precision, recall, and F1 score. Our findings show that the proposed language augmentation approaches work well across different BERT models. We demonstrate the importance of augmenting code-mixed text with language information on five different code-mixed Hindi-English downstream datasets based on sentiment analysis, hate speech detection, and emotion detection.
\end{abstract}


% \begin{highlights}
% \item Low Resource Natural Language Processing
% \item Interleaved Word-Language
% \item Adjacent Sentence-Language
% \end{highlights}
% Keywords
% Each keyword is separated by \sep
\maketitle
% Main text
\section{Introduction}\label{Introduction}
Code-mixed data, commonly observed on social networking platforms, refer to textual content that blends two or more languages. Nevertheless, natural language processing research has predominantly concentrated on monolingual data \cite{joshi2022l3cube_hindbert,joshi2022l3cube_mahacorpus}, resulting in a scarcity of code-mixed data for future investigations \cite{khanuja2020gluecos}. For instance, in India, Hindi-English code-mixing is widely prevalent on social media, owing to Hindi's status as the official language of the country \cite{patil2023comparative,nayak2021contextual}.

The growing necessity to develop models capable of effectively processing and comprehending code-mixed text has become increasingly evident, as code-mixed data presents a unique challenge for NLP research. Various pre-trained models, such as BERT and its variants, have been proposed for numerous tasks, and initial outcomes are promising. Recent studies have demonstrated that HingBERT-based models \cite{nayak-joshi-2022-l3cube}, specifically trained on code-mixed data, outperform other models for tasks involving code-mixed text categorization  \cite{orlov2022supervised,patil2023comparative}. In this study, we compare the performance of several BERT models using datasets for downstream code-mixed text classification and reaffirm that their HingBERT counterparts exhibit superior understanding and decoding capabilities for code-mixed content. Moreover, by augmenting the code-mixed text with language information, we can enhance the accuracy and efficiency of NLP tasks.

Language is a crucial component in code-mixed datasets. A word may appear in both languages yet possess distinct meanings in each. Consider the following code-mixed Hindi-English example: \textit{"\textbf{bus} ho gaya bhai, don't chatter in the \textbf{bus}, shant betho please"}. Here, the first occurrence of the word "bus" is in Hindi which means enough. However, the second occurrence of the term "bus"  is in English and refers to a vehicle. We explicitly provide this word-level language information to the model to make it easier for it to deduce its meaning.

 The prevailing state-of-the-art models, which have achieved remarkable accuracy comparable to human performance, primarily focus on mining and analyzing text from resource-rich monolingual sources. However, these models do not effectively leverage language information when it comes to code-mixed data \cite{10.1007/978-981-16-3690-5_73}. Incorporating an additional language embedding layer necessitates training the model from scratch, potentially introducing greater complexity and compromising accuracy.

To gain a deeper understanding of the diverse languages present in code-mixed texts, our research employs language identification, a process that aims to determine the language of given words. The objective of our study is to enhance BERT-based models \cite{devlin-etal-2019-bert} for code-mixed tasks by leveraging language information. By incorporating language tags, we provide additional contextual information to the language models, enabling them to better comprehend the data and achieve improved performance in downstream tasks. We employ two approaches for language augmentation named as Interleaved Word-Language method and the Adjacent Sentence-Language method. In the Interleaved Word-Language approach, language tags are inserted after each word in the sentence, while the Adjacent Sentence-Language approach appends language tags at the end of the sentence. These language tags are represented as text and are seamlessly integrated with the original input text, requiring no modifications to the model architecture. This straightforward approach is particularly relevant as it can be easily integrated with existing pre-trained BERT models.

\begin{figure}[!t]
     \centering
     \includegraphics[scale=0.35]{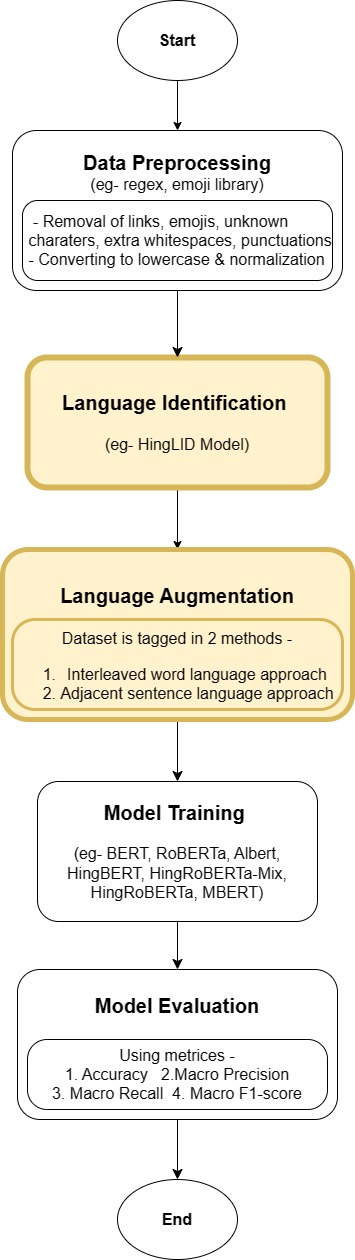}
     \caption{Proposed pipeline for language augmentation}
     \label{fig:Pipeline}
\end{figure}

In this paper, we introduce a novel pipeline for effectively processing code-mixed data by integrating the language identification model within the regular flow. The pipeline consists of pre-processing, language identification, language augmentation, and model training as depicted in Figure \ref{fig:Pipeline}. Our approach is evaluated on various downstream tasks, including sentiment analysis, emotion recognition, and hate speech detection. To conduct our evaluation, we utilize a total of five datasets, consisting of two sentiment analysis datasets, one emotions dataset, and two hate speech datasets. Through our experiments, we demonstrate that HingBERT-based models \cite{nayak-joshi-2022-l3cube} outperform other models on these datasets. Additionally, we present new analyses conducted on these datasets, leveraging the proposed language-based pipelined approach. We show that language integration improves the performance of all the BERT variants.

The main contributions of this work are as follows:
\begin{itemize}
\item We propose a language augmentation-based pipelined approach to improve the performance of code-mixed BERT models. This is the first work to integrate word-level language information into pre-trained BERT models using simple input augmentations. We do not make any architectural changes to incorporate the language information.
\item Our proposed methodology is extensively evaluated on vanilla BERT models and code-mixed HingBERT models. We demonstrate the significance of language augmentation in code-mixed text through evaluations on five datasets.
\end{itemize}
Although we present are results in the context of Hindi-English code-mixing the ideas are generic enough to be extended to any other code-mixed language. It assumes that a strong word-level LID model is available for the respective dataset.

The paper is structured as follows:  Section \ref{Literature Survey} contains details on related work in this sector. Section \ref{Methodology} describes the datasets and models used and our proposed pipeline. Section \ref{Results} includes detailed results, and Section \ref{Conclusion} contains inferences based on the results.

\begin{figure*}[!ht]
    \centering
    \includegraphics[scale=0.6]{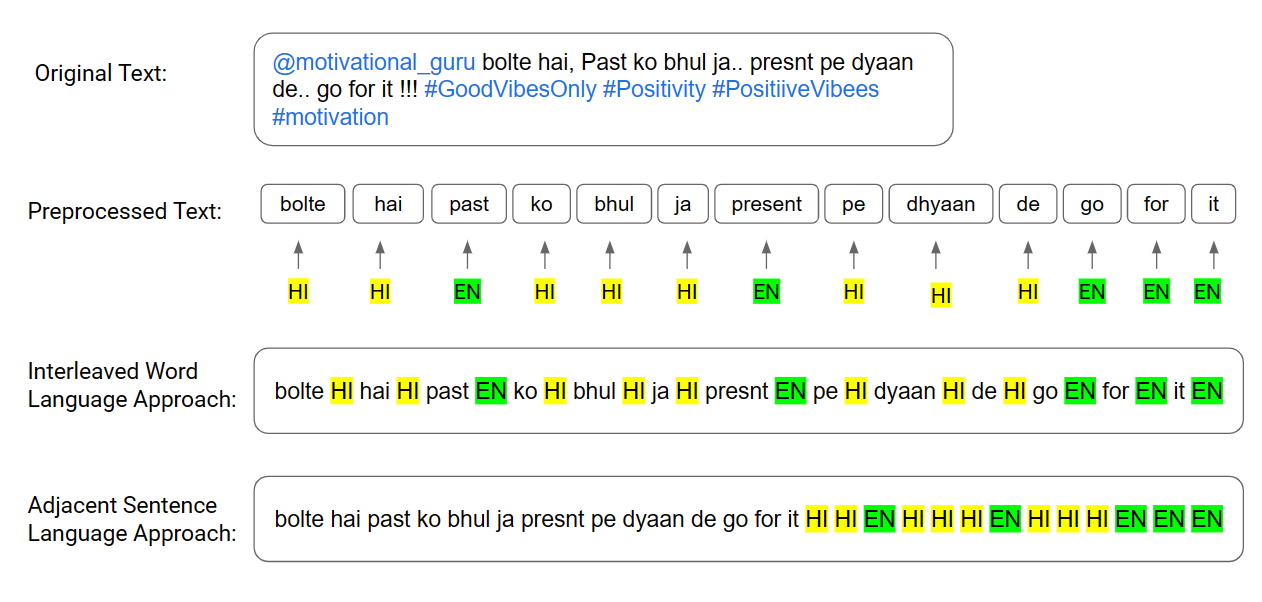}
    \caption{The demonstration of input preparation in the pipeline. The two approaches are used for language augmentation. The interleaved word language approach places the language tag after each work. The adjacent sentence language approach places the language tags after the complete sentence.}
    \label{fig:Pipeline_Example_Text}
\end{figure*}

\section{Literature Survey} \label{Literature Survey}

The code-mixed NLP has centered mainly around tasks like sentiment analysis, named entity recognition, and POS tagging. In this section, we discuss specific modifications proposed in the context of these tasks. 
Analyzing papers on sentiment analysis on code-mixed data, \cite{pradhan-et-al} use a deep learning classifier that combines a multilingual variant of RoBERTa with a sentence-level embedding from Universal Sentence Encoder. By applying both for transfer learning, they maximize the performance of the classifier. \cite{pratapa-etal-2018-word} yielded the best results by training skip-grams in the synthetic code-mixed text produced by code-mixing linguistic models. They found that existing bilingual embeddings can be improved by creating multilingual word embeddings. To leverage BERT-based models in the medical industry, \cite{alsentzer-etal-2019-publicly} used clinical embeddings to achieve better outcomes.

In \cite{alghanmi-etal-2020-combining}, AraBERT, trained on code-mixed English-Arabic models, used static word embeddings created using Word2Vec. Static word embeddings were found to be more suitable for low-resource NLP models, as AraBERT outperformed BERT models. In order to create effective sentence representations, \cite{9140343} analyzed deep contextualized models and developed an efficient sentence embedding model called SBERT-WK. Contrasting widely used word embedding strategies like Word2Vec and Glove, \cite{alatawi2020detecting} researched a deep learning-based word embedding model (BiLSTM) to tackle hate speech. 
%On the other hand, \cite{devlin-etal-2019-bert} showed that BERT is conceptually simple yet empirically powerful. It achieved new state-of-the-art results on eleven natural language processing tasks, such as pushing the GLUE score to 80.5\%

\cite{DBLP:journals/corr/abs-2107-01202}, in this work, word piece and byte-level byte pair encoding (BLBPE) were combined with BERT and RoBERTa models. Hi-En (Hindi-English) and Urdu-En (Urdu-English) databases were used, and it is worth noting that Hindi and Urdu have similar vocabularies. Various token categories, including English, Hindi, Universal, URL, hashtag, username, and named entity, were employed for token grouping. The study found that BLBPE resulted in better outcomes.

The study presented in \cite{10.1145/3380967} addresses the challenges posed by code-mixed text, specifically for Bengali-English, Hindi-English, and Telugu-English POS tagging, by utilizing widely available unlabeled data. They propose a hierarchical deep neural network-based approach, which combines CRF on top of BiLSTM, to handle long short-term dependencies at the word level and address the issue of out-of-vocabulary (OOV) words using character-level information.

To address the challenges presented by unstructured code-mixed data and enhance the performance of Named Entity Recognition (NER) on code-mixed text, \cite{9077733} proposes two methods. Firstly, it suggests utilizing word embeddings generated by word2vec and GloVe to feed them into deep learning models for word-level classification into predefined named entities. Secondly, it employs a bidirectional character-level recurrent neural network to leverage sub-word level syntactic information. This approach retains the contextual usage of words and utilizes sub-word level morphological characteristics that are not specific to any particular language, making it language-independent.

In the study conducted by \cite{singh-etal-2018-named}, various machine learning classification algorithms, including Decision Tree, LSTM, and CRF, were experimented with on a corpus for Named Entity Recognition in Hindi-English code-mixed text. The authors incorporated word, character, and lexical features in their models. The results demonstrated that both CRF and LSTM achieved the highest f1-score of 0.95. The paper discussed the reasoning behind selecting specific features for this task and compared the performance of different classification models.

The objective of the research presented in \cite{vyas-etal-2014-pos} is to create a multi-level annotated corpus of Hindi-English code-mixed text sourced from Facebook forums. The authors investigated POS tagging, normalization, back-transliteration, and language identification. They found that the accuracy of POS tagging is significantly affected by the correct identification of language and the transliteration of Hindi. The research highlights normalization and transliteration as particularly challenging tasks in English-Hindi code-mixed text and suggests that modeling language identification, normalization, transliteration, and POS tagging together can yield improved results.

\cite{lal-etal-2019-de} This work introduces a novel approach for sentiment analysis of data containing a mixture of English and Hindi code. It utilizes recurrent neural networks (RNNs) to accurately analyze sentiment information. The paper suggests modifying the attention mechanisms and expanding the training dataset to evaluate the model's performance. These modifications have the potential to improve the model's ability to capture important sentiment-related features.

In the study by \cite{joshi-etal-2016-towards}, an innovative approach is presented to enhance the accuracy of text analysis tasks, particularly when dealing with text that has a high level of noise. The proposed method focuses on acquiring sub-word level representations within the LSTM architecture, enabling better handling and extraction of meaningful information from text that exhibits excessive noise or irregularities. Furthermore, the proposed technique holds promise for broader application beyond the LSTM architecture, allowing for scalability and potential performance improvements in various text analysis tasks. Future research directions could explore the application of the proposed technique in deep neural network architectures. This research study investigates a novel approach to improve the performance of text analysis in the presence of excessive noise by utilizing sub-word level representations within the LSTM architecture. It also provides a dataset specifically designed for sentiment analysis and opens avenues for further exploration in deep neural network architectures.

The findings of \cite{DBLP:journals/corr/abs-1811-05145} contribute to the advancement of sentiment analysis in the presence of noise and lay the groundwork for future investigations. This research presents a novel approach for learning sub-word level representations within the LSTM architecture, enabling more accurate and reliable sentiment analysis in challenging text environments. It also suggests potential applications in domains involving variations of language, such as sarcasm or misinformation. Deep neural network architectures play a crucial role in addressing variations of language and textual assimilation.

The usage of LID in code-mixed tasks has not received enough attention probably because of the absence of a strong word-level LID model. Recently, a strong Hindi-English LID model named HingLID\footnote{\url{https://huggingface.co/l3cube-pune/hing-bert-lid}} was released by L3Cube allowing us to explore the methods proposed in this work.

\section{Methodology} \label{Methodology}
In this study, language identification tags in the form of interleaved word-language and adjacent sentence-language approaches are utilized to enhance the performance of models on low-resource, code-mixed Hindi-English texts for downstream natural language processing tasks. This part describes the recommended strategy, which consists of a series of steps including dataset descriptions, data preparation techniques, and model training.

\subsection{Pipeline}
To conduct the analysis on the dataset, we developed a pipeline that encompassed data preprocessing, language tagging, and prediction using Transformer-based language models. The pipeline consisted of the following components:

\textbf{Data Preprocessing:}
The dataset underwent preprocessing steps to handle links, emojis, unknown characters, extra whitespaces, and punctuation marks. Furthermore, the text was converted to lowercase to ensure data normalization. Python's regex and emoji libraries were utilized for text preprocessing. In this study, stemming and lemmatization techniques were not employed to preserve the context of the users' purpose.

\textbf{Word-level Language Identification:}
This stage of the pipeline involved language identification of each word in the text. Language identification was conducted using language identification models such as HingLID, a pre-trained language identification model provided by L3Cube. HingLID demonstrated high accuracy in identifying the language of the text within the dataset. Through the use of HingLID, we were able to distinguish between Hindi and English words. The assigned language tags were as follows:
The language tags assigned are as follows -
\begin{itemize}
    \item EN - English
    \item HI - Hindi
\end{itemize}

\textbf{Language Augmentation:}
During the preprocessing stage, all characters are converted to lowercase, and their corresponding language tags are changed to uppercase (HI/EN). After assigning the appropriate language tags to each word, we train models using two different language augmentation approaches as shown in Figure \ref{fig:Pipeline_Example_Text}.
\begin{itemize}
    \item Interleaved Word-Language approach - When using the word-lang method, we attach language tags after each individual word.\\
    
    \item Adjacent Sentence-Language approach - When using the sentence-lang method, we attach language tags of all the words at the end of the sentence.\\
    
\end{itemize}

\textbf{Model Training:}
After preprocessing and language tagging, the text is encoded and tokenized using BERT tokenizer. The data is then fed as input to BERT-based models for respective downstream tasks. The models are trained on three types of data: original data without language tags, data augmented using the interleaved word-language approach, and data augmented using the adjacent sentence-language approach. For hyperparameter tuning, WandB, a tool for hyperparameter optimization, is utilized. The learning rate is adjusted within the range of 1e-6 to 1e-4, the epoch range is set from 1 to 7, and the batch size ranges from 32 to 64.

\textbf{Evaluation:}
To assess the performance of the pipeline, the following evaluation metrics are employed: accuracy, macro precision, macro recall, and macro F1 score. Accuracy represents the percentage of correctly classified samples in the dataset, while the F1 score is the harmonic mean of precision and recall. Precision measures the proportion of true positives among all positive predictions, while recall measures the proportion of true positives among all actual positives. A comparative analysis is conducted between BERT-based models and their Hing counterparts across the three types of data: original data without language tags, data augmented using the interleaved word-language technique, and data enhanced using the adjacent sentence-language strategy.

\subsection{Datasets}
\begin{figure*}[!ht]
    \centering
    \includegraphics[scale=0.4]{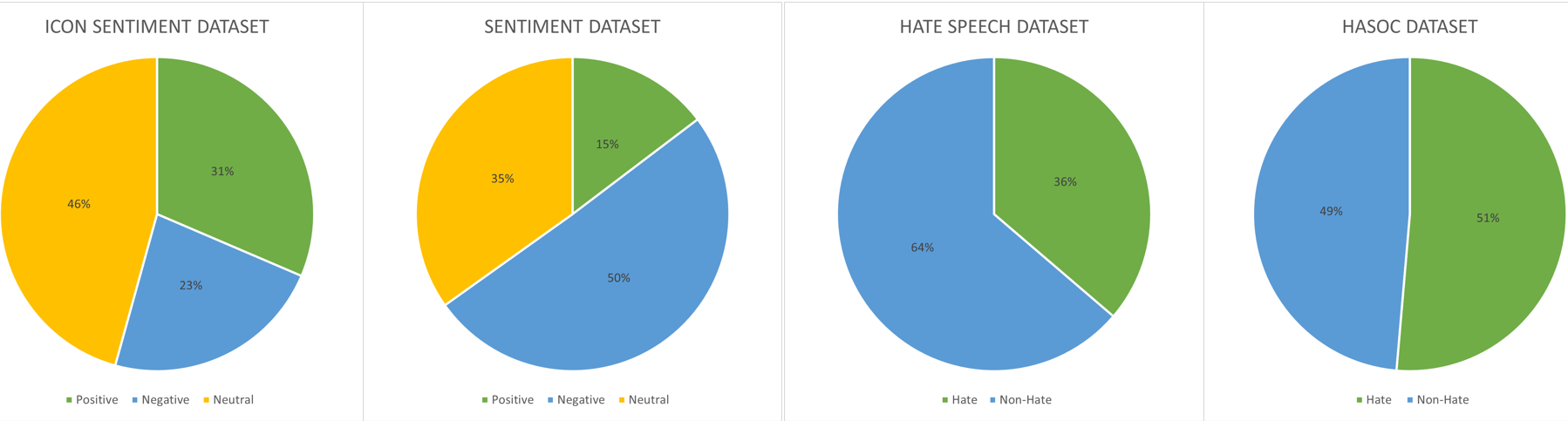}
    \caption{Icon Sentiment, Sentiment, Hate Speech and HASOC Dataset}
    \label{fig:Datasets}
\end{figure*}

\begin{figure}[!ht]
     \centering
     \includegraphics[scale=0.5]{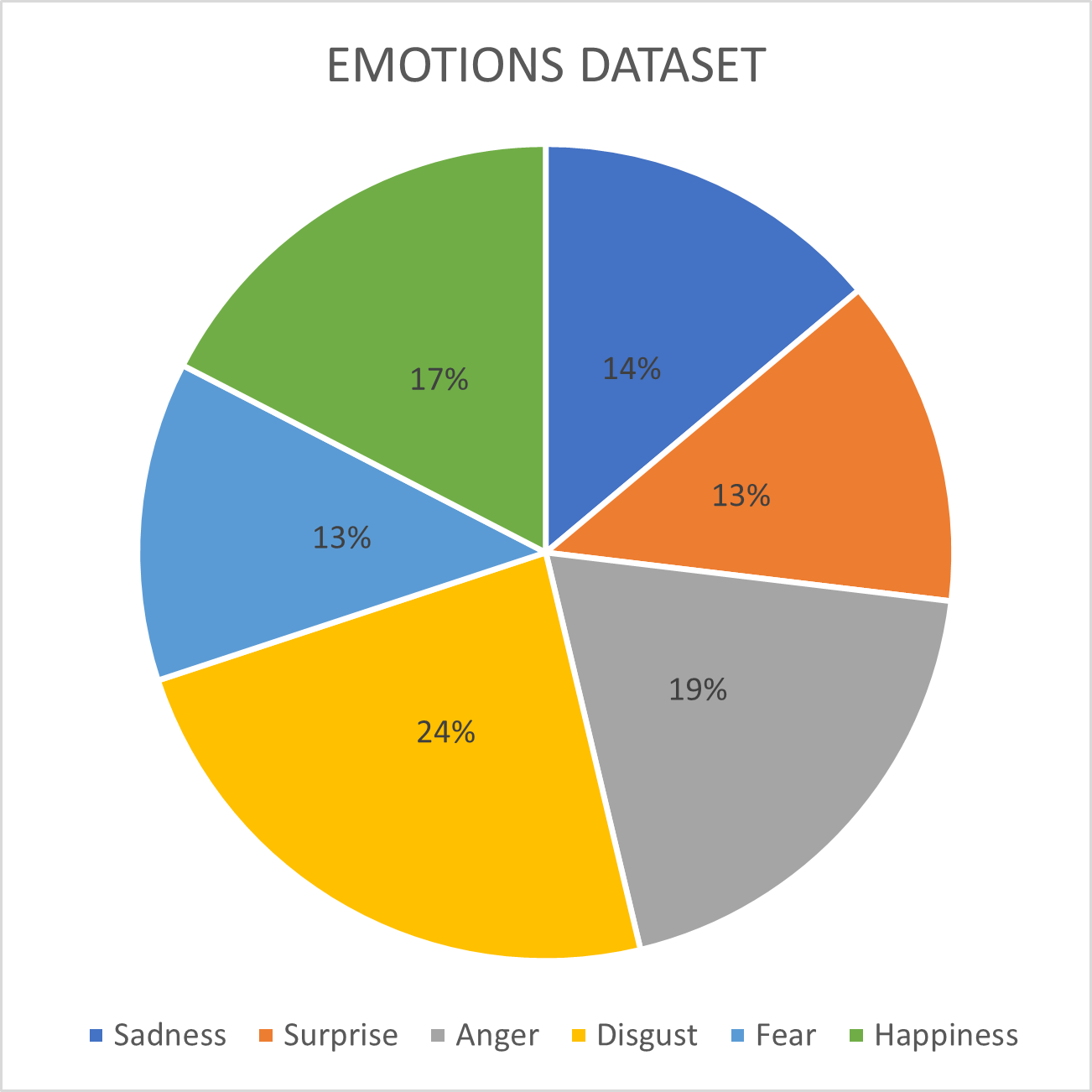}
     \caption{Emotions Dataset}
     \label{fig:Emotions_Dataset}
\end{figure}

The label statistics of the data used in this work are shown in Figure \ref{fig:Datasets} and Figure \ref{fig:Emotions_Dataset}. These datasets are described in the subsequent sub-sections.

\subsubsection{Icon Sentiment Dataset}
The Icon Dataset is derived from the Icon shared task \cite{Patra2018SentimentAO}. It consists of tweets that contain a mixture of Hindi and English codes. The dataset comprises 5,525 test instances and 12,936 training instances. The tweets exhibit three primary emotions: positive, negative, and neutral. The baseline performance for the Icon shared task was established with a baseline F1 score of 0.331. Among the participating institutions, IIIT-NBP achieved the highest F1 score of 0.569, highlighting the effectiveness of their approach. Importantly, the dataset includes linguistic information that offers valuable insights for research purposes. \\

\subsubsection{Sentiment Dataset}
This dataset, described in \cite{DBLP:journals/corr/PrabhuJSV16}, aims to classify tweet sentiments into three categories: negative, neutral, and positive. The dataset contains a total of 3,879 tweets. The Subword-LSTM approach achieves an F1 score of 0.658 and a maximum accuracy score of 69.7\%. To evaluate the model's performance, the dataset is divided into training and testing sets using a 70:15:15 ratio. \\

\subsubsection{Hate Speech Dataset}
This code-mixed Hindi-English dataset was compiled from the research conducted in \cite{bohra-etal-2018-dataset}, which focused on hate speech detection. The dataset is categorized into two classes: hate speech and regular speech. It comprises a total of 4,575 tweets, with 1,661 examples of hate speech and 2,914 examples of normal speech. When considering all features, Support Vector Machines achieved the highest accuracy result of 71.7\%. The dataset is divided into three parts in a ratio of 70:15:15, where 70\% of the data is used for training, 15\% for validation, and 15\% for testing. Language recognition tagging was performed using the Hing-LID model. Preprocessing involved removing emojis, links, and punctuation, so that only alphanumeric characters were retained.\\

\subsubsection{HASOC Dataset}
This investigation utilized data from the HASOC Shared Task 2022 as the source database for the analysis. The data was divided into two categories: hateful speech and non-hateful speech. The dataset consists of a total of 4,864 tweets, with 2,514 classified as hate speech and 2,390 as non-hate speech. Among the participating teams, the NLPLAB-ISI team achieved the highest F1 score of 0.708. To prepare the dataset for training and evaluation, it was divided into three portions using a ratio of 70:15:15. Specifically, 70\% of the dataset was used for training, 15\% for validation, and the remaining 15\% for testing.\\

\subsubsection{Emotions Dataset}
This dataset was obtained from research \cite{https://doi.org/10.48550/arxiv.2102.09943} focused on emotion detection. The tweets in the dataset are categorized into six emotions. There is a total of 151,311 messages, with the following distribution: 26,364 tweets conveying happiness, 21,024 expressing sadness, 29,306 indicating anger, 19,138 conveying fear, 35,797 representing disgust, and 19,682 indicating surprise. The BERT model achieved the highest accuracy value of approximately 71.43\%. The dataset is divided into three parts: training, validation, and testing, with a ratio of 70:15:15, respectively. Language recognition tagging was performed using the Hing-LID model. During preprocessing, emojis, links, and punctuation were removed to retain only alphanumeric characters.

\begin{table}[!ht]
\caption{Dataset Train-Test-Validation Split}\label{tab:DatasetsSplit}
\begin{tabular*}{\tblwidth}{@{}C@{}L@{}C@{}C@{}C@{}C@{}C}
\toprule
Sr. & Dataset & Total & Train & Test & Eval & Labels\\ % Table header row
No. & Name & size & Size & Size & Size & \\
\midrule
1 & Icon Dataset & 18461 & 12936 & 5525 & 5525 & 3\\
2 & Sentiment Dataset & 3879 & 2,715 & 582 & 582 & 3\\
3 & Hatespeech Dataset &4578 & 3204 & 687 & 687 & 2 \\
4 & HASOC Dataset & 4864 & 3404 & 730 & 730 & 2 \\
5 & Emotions Dataset & 151311 & 105917 & 22697 & 22697 & 6\\
\bottomrule
\end{tabular*}
\end{table}

% \begin{figure*}[!ht]
%     \centering
%     \includegraphics[scale=0.4]{imgcomp.png}
%     \caption{Icon Sentiment, Sentiment, Hate Speech and HASOC Dataset}
%     \label{fig:Datasets}
% \end{figure*}

% \begin{figure}[!ht]
%      \centering
%      \includegraphics[scale=0.5]{EMOTIONS.png}
%      \caption{Emotions Dataset}
%      \label{fig: Emotions Dataset}
% \end{figure}

% \begin{table}[!ht]
% \caption{Dataset Train-Test-Validation Split}\label{tab:DatasetsSplit}
% \begin{tabular*}{\tblwidth}{@{}C@{}L@{}C@{}C@{}C@{}C@{}C}
% \toprule
% Sr. & Dataset & Total & Train & Test & Eval & Labels\\ % Table header row
% No. & Name & size & Size & Size & Size & \\
% \midrule
% 1 & Icon Dataset & 18461 & 12936 & 5525 & 5525 & 3\\
% 2 & Sentiment Dataset & 3879 & 2,715 & 582 & 582 & 3\\
% 3 & Hatespeech Dataset &4578 & 3204 & 687 & 687 & 2 \\
% 4 & HASOC Dataset & 4864 & 3404 & 730 & 730 & 2 \\
% 5 & Emotions Dataset & 151311 & 105917 & 22697 & 22697 & 6\\
% \bottomrule
% \end{tabular*}
% \end{table}

Table \ref{tab:DatasetsSplit}  shows the size of the corpus, its train, test, and validation splits, as well as the number of labels associated with each dataset.
% \FloatBarrier

\subsection{Language Tags Analysis}

% \begin{table}[!ht]
% \caption{LID Analysis}\label{tab:LIDAnalysis}
% \begin{tabular*}{\tblwidth}{@{}L@{}C@{}C@{}C@{}C@{}C}
% \toprule
% Dataset. & Hindi & English & Hindi & English & Ratio \\ % Table header row
%  &  &  & tokens & tokens & \\
%  &  &  & per & per & \\
%  &  &  & Sentence & Sentence & \\
% \midrule
% ICON & 68159 & 67900 & 6.81 & 6.78 & 0.53 \\
% Sentiment & 31770 & 11318 & 8.19 & 2.92 & 0.72 \\
% Hate Speech & 69089 & 21240 & 15.09 & 4.64 & 0.74 \\
% HASOC & 37910 & 59444 & 7.72 & 12.11 & 0.44 \\
% Emotions & 1944214 & 515550 & 12.85 & 3.41 & 0.77 \\
% \bottomrule
% \end{tabular*}
% \end{table}

The Table \ref{tab:LIDAnalysis} presents a summary of the analysis conducted on each dataset using LID (Language Identification). It includes the total count of Hindi and English tokens observed across all datasets, as well as the average count of Hindi and English tokens per sentence. Furthermore, the ratio of Hindi tokens to the total number of tokens in each sentence is calculated, providing valuable insights into the code-mixed dataset.

\begin{table}[!ht]
\caption{LID Analysis}\label{tab:LIDAnalysis}
\begin{tabular*}{\tblwidth}{@{}L@{}C@{}C@{}C@{}C@{}C}
\toprule
Dataset. & Hindi & English & Hindi & English & Ratio \\ % Table header row
 &  &  & tokens & tokens & \\
 &  &  & per & per & \\
 &  &  & Sentence & Sentence & \\
\midrule
ICON & 68159 & 67900 & 6.81 & 6.78 & 0.53 \\
Sentiment & 31770 & 11318 & 8.19 & 2.92 & 0.72 \\
Hate Speech & 69089 & 21240 & 15.09 & 4.64 & 0.74 \\
HASOC & 37910 & 59444 & 7.72 & 12.11 & 0.44 \\
Emotions & 1944214 & 515550 & 12.85 & 3.41 & 0.77 \\
\bottomrule
\end{tabular*}
\end{table}

\subsection{Models}
We used the following models in our research:\\

\begin{itemize}
    \item ALBERT\footnote{\url{https://huggingface.co/albert-base-v2}}: A Lite BERT (ALBERT) model introduced in \cite{DBLP:journals/corr/abs-1909-11942} for self-supervised learning of language representations. This model shares the same architectural foundation as BERT, but its main objective was to enhance BERT's performance by implementing various design choices that reduce the number of training parameters, making it more memory and time efficient.
    \item BERT\footnote{\url{https://huggingface.co/bert-base-cased}}: \cite{DBLP:journals/corr/abs-1810-04805} proposed the BERT model that uses self-supervised transformers and was pre-trained on a large corpus of multilingual data, which includes the Toronto Book Corpus and Wikipedia. This pre-training was conducted using raw text and did not involve human labeling. The training process included masked language modeling and the next sentence prediction task.
    \item RoBERTa\footnote{\url{https://huggingface.co/roberta-base}}: RoBERTa, proposed in \cite{DBLP:journals/corr/abs-1907-11692}, which stands for Robustly Optimized BERT Pretraining Approach, is a model based on Google's BERT model from 2018. It differs from BERT in that it removes the next-sentence pretraining step and uses much larger batches during training, resulting in faster and more effective training.
    \item Multilingual BERT\footnote{\url{https://huggingface.co/bert-base-multilingual-cased}}: The mBERT model proposed in \cite{https://doi.org/10.48550/arxiv.1810.04805} is a version of BERT that was trained on a massive multilingual dataset, which includes 104 different languages.
    \item HingBERT\footnote{\url{https://huggingface.co/l3cube-pune/hing-bert}}: A BERT model called HingBERT was created by continued pre-training on Hindi-English code-mixed Roman text, as mentioned in \cite{nayak-joshi-2022-l3cube}. This model is an upgraded version of the mBERT model and was trained on the L3Cube-HingCorpus, which consisted of 1.04 billion tokens from 52.93 million Twitter phrases.
    \item HingRoBERTa\footnote{\url{https://huggingface.co/l3cube-pune/hing-roberta}}: The HingRoBERTa proposed in \cite{nayak-joshi-2022-l3cube} is a RoBERTa model that has been fine-tuned on a mix of Hindi and English codes in Roman text. The fine-tuning process was conducted using the Roman version of L3Cube-HingCorpus.
    \item HingRoBERTa-Mixed\footnote{\url{https://huggingface.co/l3cube-pune/hing-roberta-mixed}}: The HingRoBERTa-Mixed proposed in \cite{nayak-joshi-2022-l3cube} is a BERT model that combines Hindi and English codes in both Roman and Devanagari script. It is an XLM-RoBERTa model that has been fine-tuned on a mix of Roman and Devanagari code-mixed text from L3Cube-HingCorpus.
\end{itemize}

\section{Results} \label{Results}

The models are evaluated using three configurations: interleaved word-language, adjacent sentence-language, and without linguistic information. The F1 score, a metric that considers both precision and recall, measures the models' performance and offers valuable insights into the performance variations, highlighting the significance of choosing the appropriate configuration for optimal results. The results presented in the paper are comprehensive, revealing the impact of these approaches on the model's effectiveness across the datasets. The general trend observed across all datasets is that the Hing-based BERT models outperform the vanilla BERT-based models, and the proposed pipeline produces better results than utilizing data without language information.
% \newpage

\begin{figure*}[!ht]
    \centering
    \includegraphics[scale=0.22]{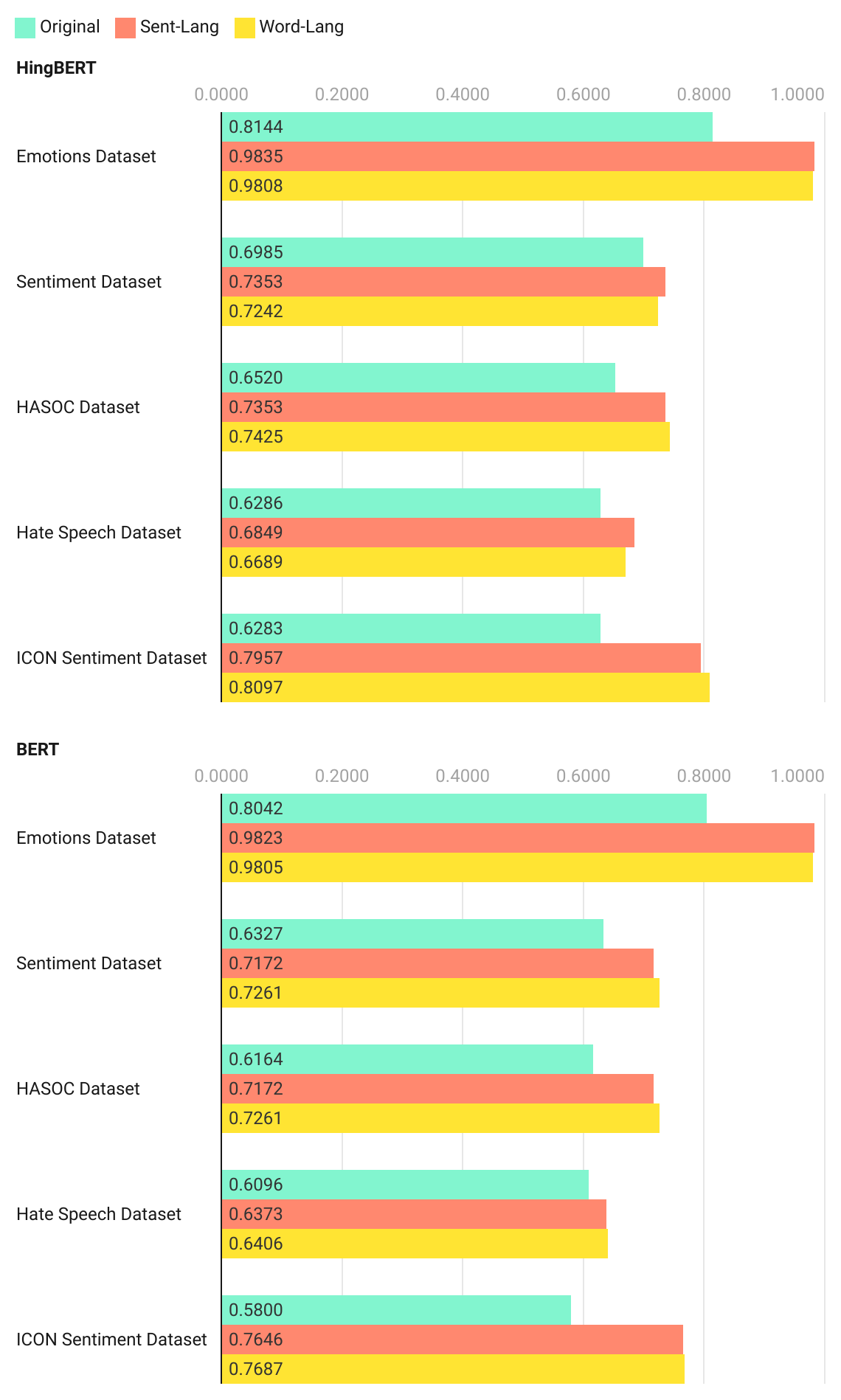}
    \caption{Results: Comparison of vanilla, word-lang, and sent-lang approaches using HingBERT and BERT}
    \label{fig:Results}
\end{figure*}

Figure \ref{fig:Results} shows the comparison of interleaved word-language, adjacent sentence-language, and without language tags on HingBERT and BERT. Comparative F1 score graphs are shown, and the outcomes highlight HingBERT's supremacy in comprehending and handling code-mixed data by outperforming BERT across all techniques. The comprehensive results for the remaining models are presented in the following tables.

\begin{table*}[!ht]
\caption{Results displaying F1 Score}\label{tab:Results}
\begin{tabular*}{\tblwidth}{@{}C@{}L@{}C@{}C@{}C@{}C@{}C}
\toprule
Models & Approach & Icon & Sentiment & Hate Speech & HASOC & Emotions\\ % Table header row
\midrule
\textbf{} & Interleaved Word-Language & 0.79573 & 0.72416 & 0.66892 & \textbf{0.74245} & 0.98080\\
HingBERT & Adjacent Sentence-Language & 0.80974 & 0.73531 & 0.68492 & 0.73531 & 0.98353\\
\textbf{} & Without Language Tags & 0.62825 & 0.69845 & 0.62860 & 0.65204 & 0.81437\\
\midrule
\textbf{} & Interleaved Word-Language & 0.81553 & 0.75267 & \textbf{0.69994} & 0.72087 & 0.98425\\
HingRoBERTa & Adjacent Sentence-Language & \textbf{0.81685} & 0.74359 & 0.64628 & 0.73437 & 0.98362\\
\textbf{} & Without Language Tags & 0.65222 & 0.75515 & 0.66994 & 0.73270 & 0.98195\\
\midrule
\textbf{} & Interleaved Word-Language & 0.80100 & 0.73773 & 0.67182 & 0.75454 & \textbf{0.98447}\\
HingRoBERTa-Mixed & Adjacent Sentence-Language & 0.80322 & 0.73137 & 0.66121 & 0.73295 & 0.98357\\
\textbf{} & Without Language Tags & 0.65288 & \textbf{0.76160} & 0.65990 & 0.71087 & 0.98285\\
\midrule
\textbf{} & Interleaved Word-Language & 0.73172 & 0.65767 & 0.62268 & 0.68583 & 0.97581\\
Multilingual BERT & Adjacent Sentence-Language & 0.73744 & 0.67032 & 0.61958 & 0.72328 & 0.98233\\
\textbf{} & Without Language Tags & 0.56593 & 0.65464 & 0.54588 & 0.67193 & 0.97705\\
\midrule
\textbf{} & Interleaved Word-Language & 0.71394 & 0.58346 & 0.63354 & 0.68985 & 0.97316\\
AlBERT & Adjacent Sentence-Language & 0.73243 & 0.57939 & 0.59515 & 0.70455 & 0.97255\\
\textbf{} & Without Language Tags & 0.54280 & 0.61211 & 0.61399 & 0.59617 & 0.95540\\
\midrule
\textbf{} & Interleaved Word-Language & 0.76872 & 0.72609 & 0.64063 & 0.72609 & 0.98047\\
BERT & Adjacent Sentence-Language & 0.76463 & 0.71724 & 0.63730 & 0.71724 & 0.98228\\ 
\textbf{} & Without Language Tags & 0.58004 & 0.63273 & 0.60958 & 0.61643 & 0.80423\\
\midrule
\textbf{} & Interleaved Word-Language & 0.73566 & 0.63912 & 0.62285 & 0.69897 & 0.98017\\
RoBERTa & Adjacent Sentence-Language & 0.75555 & 0.66375 & 0.59617 & 0.69610 & 0.98149\\
\textbf{} & Without Language Tags & 0.57978 & 0.66237 & 0.63983 & 0.66259 & 0.98001\\
\bottomrule
\end{tabular*}
\end{table*}

Table \ref{tab:Results} displays the F1 score results for all the datasets.

\begin{itemize}
    \item \textbf{Icon Dataset:} The HingRoBERTa model achieves the best result for sentiment analysis on the Icon Dataset using the Adjacent Sentence-Language approach. There is a significant increase in the F1 score for the HingRoBERTa model when using the word-lang (0.163) and sentence-lang (0.164) approaches compared to using data without language information.
    \item \textbf{Sentiment Dataset:} We observe improvement with the addition of language tags for all the models except two HingRoBERTa models. The HingRoBERTa and HingRoBERTa-Mixed models do not benefit from language tags for this sentiment analysis task. 
    \item \textbf{Hate Speech Dataset:} On this hate speech identification dataset, the HingRoBERTa model along with Interleaved Word-Language technique provides the best results. We see consistent improvements with the language augmentation approaches.
    \item \textbf{HASOC Dataset:} The HingRoBERTa-Mixed model using the Interleaved Word-Language approach achieves the best score on the HASOC shared task dataset. We observed a 0.11 increase in the F1 score for the BERT model when incorporating language information using the word-lang technique for this hate speech detection downstream task.
    \item \textbf{Emotions Dataset:} For emotions analysis on the Emotions Dataset, the Interleaved Word-Language technique yields the best results when utilizing the HingRoBERTa-Mixed model. Both the HingBERT and BERT models show significant improvement, with an approximate F1 score increase of 0.18.
\end{itemize}

% \newpage
\FloatBarrier

\section{Conclusion} \label{Conclusion}
In this work, we show the significance of word-level language identification for improving the code-mixed models. We introduce language identification and language augmentation in the regular BERT flow to enhance the code-mixed (BERT) models. The idea is to integrate word-level language information into the regular BERT flow. This information allows the model to better understand the ambiguous words in the code-mixed text.
Specifically, we propose two simple language augmentation techniques like Adjacent Sentence-Language and Interleaved Word-Language. These approaches have been shown to enhance the overall F1 score and improve the performance of downstream tasks. Language tags provide specific information about the language used within a text, enabling improved comprehension and processing of code-mixed data. We show that, by assigning appropriate language labels to individual words, models can more accurately distinguish between different languages present in the text. We show the benefits of the proposed pipeline across 5 different downstream tasks including sentiment analysis, hate speech analysis, and emotion analysis. The different BERT models considered in this work are base BERT, AlBERT, RoBERTa, mBERT, HingBERT, HingRoBERTa, and HingRoBERTa-Mixed.
%Models are able to adapt to language transitions and improve their performance thanks to the presence of language tags that explicitly indicate the boundaries of different languages.
%As a result, processing code-mixed data achieves greater success when language tags are employed, particularly with the use of Adjacent Sentence-Language and Interleaved Word-Language techniques. These approaches facilitate better language identification and comprehension of mixed-language contexts, ultimately enhancing performance on a range of downstream tasks.

\section*{Acknowledgements}
This work was done under the L3Cube Pune mentorship
program. The problem statement and ideas presented in this work originated from L3Cube and its mentors. We would like to express our gratitude towards
our mentors at L3Cube for their continuous support and
encouragement.

% \section{}\label{}
% Use the style of numbering in square brackets.
% If nothing is used, default style will be taken.
%\begin{enumerate}[a)]
%\item 
%\item 
%\item 
%\end{enumerate}  

%\begin{itemize}
%\item 
%\item 
%\item 
%\end{itemize}  

%\begin{description}
%\item[]
%\item[] 
%\item[] 
%\end{description}  

% Figure
% \begin{figure}[<options>]
% 	\centering
% 		\includegraphics[<options>]{}
% 	  \caption{}\label{fig1}
% \end{figure}

% \begin{table}[<options>]
% \begin{tabular*}{\tblwidth}{@{}LL@{}}
% \toprule
% \midrule
%  & \\
%  & \\
%  & \\
%  & \\
% \bottomrule
% \end{tabular*}
% \end{table}

% Uncomment and use as the case may be
%\begin{theorem} 
%\end{theorem}

% Uncomment and use as the case may be
%\begin{lemma} 
%\end{lemma}


% \section{}\label{}


% \bibliographystyle{cas-model2-names}

% \bibliography{cas-refs}

% \newpage
\appendix
\onecolumn
\begin{table}[!ht]
\section{Appendix}\label{Appendix}
In this section, we present additional results using metrics: accuracy, macro precision, macro recall, macro f1 score with hyperparameters used by each model with each approach according to our proposed methodology across all 5 datasets.
\subsection{Icon Sentiment Dataset}
\caption{Icon Results}\label{tab:IconResults}
\begin{tabular*}{\tblwidth}{@{}C@{}C@{}C@{}C@{}C@{}C@{}C@{}C@{}C}
\toprule
Models & Approach & Batch Size & Learning Rate & Epochs & Accuracy & Precision & Recall & F1 Score\\ % Table header row
\midrule
 & Interleaved Word-Language & 64 & 1.73E-05 & 3 & 0.79879 & 0.79407 & 0.80037 & 0.79573\\
HingBERT & Adjacent Sentence-Language & 64 & 5.47E-05 & 2 & 0.81185 & 0.80872 & 0.81669 & 0.80974\\
 & Without Language Tags & 64 & 9.20E-05 & 1 & 0.64633 & 0.64530 & 0.62069 & 0.62825\\
\midrule
 & Interleaved Word-Language & 64 & 2.93E-05 & 3 & 0.81989 & 0.81451 & 0.81677 & 0.81553\\
HingRoBERTa & Adjacent Sentence-Language & 64 & 4.93E-05 & 2 & 0.82056 & 0.81520 & 0.81939 & 0.81685\\
 & Without Language Tags & 32 & 1.10E-05 & 4 & 0.66407 & 0.65765 & 0.64793 & 0.65222\\
\midrule
 & Interleaved Word-Language & 64 & 3.95E-05 & 5 & 0.80583 & 0.80018 & 0.80198 & 0.80100\\
HingRoBERTa-Mixed & Adjacent Sentence-Language & 64 & 3.48E-05 & 5 & 0.80716 & 0.80162 & 0.80574 & 0.80322\\
 & Without Language Tags & 64 & 4.84E-05 & 3 & 0.66425 & 0.65711 & 0.64940 & 0.65288\\
\midrule
 & Interleaved Word-Language & 32 & 3.97E-05 & 5 & 0.73853 & 0.73131 & 0.73218 & 0.73172\\
Multilingual BERT & Adjacent Sentence-Language & 32 & 1.52E-05 & 4 & 0.74222 & 0.73616 & 0.74011 & 0.73744\\
 & Without Language Tags & 32 & 7.25E-05 & 3 & 0.58643 & 0.57110 & 0.56441 & 0.56593\\
\midrule
 & Interleaved Word-Language & 64 & 4.08E-05 & 3 & 0.71510 & 0.71853 & 0.72433 & 0.71394\\
AlBERT & Adjacent Sentence-Language & 32 & 7.03E-05 & 7 & 0.73586 & 0.73167 & 0.73726 & 0.73243\\
 & Without Language Tags & 64 & 6.18E-05 & 4 & 0.57756 & 0.56233 & 0.53655 & 0.54280\\
\midrule
 & Interleaved Word-Language & 32 & 5.68E-05 & 3 & 0.77101 & 0.76852 & 0.77573 & 0.76872\\
BERT & Adjacent Sentence-Language & 32 & 6.26E-05 & 2 & 0.77034 & 0.76391 & 0.76550 & 0.76463\\
 & Without Language Tags & 32 & 5.81E-05 & 2 & 0.61104 & 0.60080 & 0.57166 & 0.58004\\
\midrule
 & Interleaved Word-Language & 64 & 6.73E-05 & 3 & 0.73920 & 0.73475 & 0.74025 & 0.73566\\
RoBERTa & Adjacent Sentence-Language & 64 & 6.62E-05 & 7 & 0.76197 & 0.75530 & 0.75581 & 0.75555\\
 & Without Language Tags & 32 & 7.36E-06 & 4 & 0.58914 & 0.57916 & 0.58834 & 0.57978\\
\bottomrule
\end{tabular*}
\end{table}

\begin{table}[!ht]
\subsection{Sentiment Dataset}
\caption{Sentiment Results}\label{tab:SentimentResults}
\begin{tabular*}{\tblwidth}{@{}C@{}C@{}C@{}C@{}C@{}C@{}C@{}C@{}C}
\toprule
Models & Approach & Batch Size & Learning Rate & Epochs & Accuracy & Precision & Recall & F1 Score\\ % Table header row
\midrule
 & Interleaved Word-Language & 32 & 9.66E-05 & 3 & 0.75086 & 0.74423 & 0.71731 & 0.72416\\
HingBERT & Adjacent Sentence-Language & 64 & 8.73E-05 & 7 & 0.73577 & 0.73547 & 0.73521 & 0.73531\\
 & Without Language Tags & 32 & 2.26E-05 & 4 & 0.67471 & 0.68192 & 0.66963 & 0.69845\\
\midrule
 & Interleaved Word-Language & 32 & 5.61E-05 & 2 & 0.77491 & 0.77257 & 0.74053 & 0.75267\\
HingRoBERTa & Adjacent Sentence-Language & 64 & 9.58E-05 & 5 & 0.76289 & 0.74431 & 0.74295 & 0.74359\\
 & Without Language Tags & 32 & 1.10E-05 & 4 & 0.74705 & 0.75763 & 0.73833 & 0.75515\\
\midrule
 & Interleaved Word-Language & 64 & 7.51E-05 & 2 & 0.76117 & 0.75660 & 0.72490 & 0.73773\\
HingRoBERTa-Mixed & Adjacent Sentence-Language & 32 & 6.43E-05 & 4 & 0.75258 & 0.74078 & 0.72503 & 0.73137\\
 & Without Language Tags & 64 & 8.30E-05 & 2 & 0.74898 & 0.74997 & 0.74937 & 0.76160\\
\midrule
 & Interleaved Word-Language & 32 & 6.85E-05 & 6 & 0.68900 & 0.66159 & 0.65507 & 0.65767\\
Multilingual BERT & Adjacent Sentence-Language & 32 & 6.40E-05 & 4 & 0.69931 & 0.68227 & 0.66200 & 0.67032\\
 & Without Language Tags & 32 & 7.95E-05 & 5 & 0.62456 & 0.62454 & 0.62471 & 0.65464\\
\midrule
 & Interleaved Word-Language & 32 & 3.45E-05 & 7 & 0.62543 & 0.60086 & 0.57412 & 0.58346\\
AlBERT & Adjacent Sentence-Language & 64 & 3.80E-05 & 7 & 0.62027 & 0.60032 & 0.56735 & 0.57939\\
 & Without Language Tags & 64 & 5.37E-05 & 5 & 0.57018 & 0.57840 & 0.56850 & 0.61211\\
\midrule
 & Interleaved Word-Language & 64 & 9.20E-05 & 5 & 0.72629 & 0.72603 & 0.72623 & 0.72609\\
BERT & Adjacent Sentence-Language & 64 & 2.58E-05 & 7 & 0.71816 & 0.71812 & 0.71704 & 0.71724\\
 & Without Language Tags & 32 & 6.42E-05 & 4 & 0.59549 & 0.60492 & 0.59056 & 0.63273\\
\midrule
 & Interleaved Word-Language & 64 & 5.69E-05 & 6 & 0.66323 & 0.63110 & 0.65214 & 0.63912\\
RoBERTa & Adjacent Sentence-Language & 64 & 4.63E-05 & 7 & 0.68900 & 0.67350 & 0.65653 & 0.66375\\
 & Without Language Tags & 32 & 7.62E-05 & 5 & 0.62381 & 0.63988 & 0.61984 & 0.66237\\
\bottomrule
\end{tabular*}
\end{table}

\begin{table}[!ht]
\subsection{Hate Speech Dataset}
\caption{Hate Speech Result}\label{tab:HateSpeechResult}
\begin{tabular*}{\tblwidth}{@{}C@{}C@{}C@{}C@{}C@{}C@{}C@{}C@{}C}
\toprule
Models & Approach & Batch Size & Learning Rate & Epochs & Accuracy & Precision & Recall & F1 Score\\ % Table header row
\midrule
 & Interleaved Word-Language & 64 & 8.49E-05 & 3 & 0.69869 & 0.66909 & 0.66876 & 0.66892\\
HingBERT & Adjacent Sentence-Language & 32 & 1.48E-05 & 4 & 0.71325 & 0.68510 & 0.68474 & 0.68492\\
 & Without Language Tags & 32 & 2.66E-05 & 2 & 0.69432 & 0.65713 & 0.62398 & 0.62860\\
\midrule
 & Interleaved Word-Language & 32 & 3.12E-05 & 3 & 0.73071 & 0.70349 & 0.69724 & 0.69994\\
HingRoBERTa & Adjacent Sentence-Language & 32 & 2.88E-05 & 4 & 0.70888 & 0.68132 & 0.64037 & 0.64628\\
 & Without Language Tags & 64 & 4.14E-05 & 5 & 0.71470 & 0.68239 & 0.66425 & 0.66994\\
\midrule
 & Interleaved Word-Language & 32 & 4.42E-05 & 4 & 0.69287 & 0.66914 & 0.67763 & 0.67182\\
HingRoBERTa-Mixed & Adjacent Sentence-Language & 64 & 3.69E-05 & 5 & 0.68850 & 0.65992 & 0.66282 & 0.66121\\
 & Without Language Tags & 32 & 8.32E-05 & 5 & 0.71761 & 0.68863 & 0.65266 & 0.65990\\
\midrule
 & Interleaved Word-Language & 64 & 6.42E-05 & 5 & 0.65793 & 0.62332 & 0.62211 & 0.62268\\
Multilingual BERT & Adjacent Sentence-Language & 64 & 5.18E-05 & 5 & 0.64483 & 0.61803 & 0.62347 & 0.61958\\
 & Without Language Tags & 64 & 3.86E-05 & 5 & 0.68850 & 0.68805 & 0.57017 & 0.54588\\
\midrule
 & Interleaved Word-Language & 64 & 4.43E-05 & 5 & 0.69141 & 0.65572 & 0.62883 & 0.63354\\
AlBERT & Adjacent Sentence-Language & 64 & 3.30E-05 & 5 & 0.69141 & 0.66688 & 0.59926 & 0.59515\\
 & Without Language Tags & 64 & 9.38E-05 & 5 & 0.70742 & 0.69097 & 0.61426 & 0.61399\\
\midrule
 & Interleaved Word-Language & 32 & 8.03E-05 & 4 & 0.68268 & 0.64700 & 0.63736 & 0.64063\\
BERT & Adjacent Sentence-Language & 64 & 4.90E-05 & 5 & 0.67249 & 0.63867 & 0.63619 & 0.63730\\
 & Without Language Tags & 32 & 3.44E-05 & 3 & 0.71325 & 0.71431 & 0.61279 & 0.60958\\
\midrule
 & Interleaved Word-Language & 64 & 1.97E-05 & 5 & 0.69287 & 0.65940 & 0.61946 & 0.62285\\
RoBERTa & Adjacent Sentence-Language & 64 & 3.22E-05 & 5 & 0.65357 & 0.60801 & 0.59396 & 0.59617\\
 & Without Language Tags & 64 & 1.97E-05 & 4 & 0.72344 & 0.71549 & 0.63539 & 0.63983\\
\bottomrule
\end{tabular*}
\end{table}

\begin{table}[!ht]
\subsection{HASOC Dataset}
\caption{HASOC Results}\label{tab:HASOCResults}
\begin{tabular*}{\tblwidth}{@{}C@{}C@{}C@{}C@{}C@{}C@{}C@{}C@{}C}
\toprule
Models & Approach & Batch Size & Learning Rate & Epochs & Accuracy & Precision & Recall & F1 Score\\ % Table header row
\midrule
 & Interleaved Word-Language & 64 & 8.43E-05 & 6 & 0.74255 & 0.74250 & 0.74280 & 0.74245\\
HingBERT & Adjacent Sentence-Language & 64 & 8.73E-05 & 7 & 0.73577 & 0.73547 & 0.73521 & 0.73531\\
 & Without Language Tags & 64 & 4.75E-05 & 3 & 0.65205 & 0.65205 & 0.65204 & 0.65204\\
\midrule
 & Interleaved Word-Language & 64 & 4.12E-05 & 7 & 0.72087 & 0.72176 & 0.72176 & 0.72087\\
HingRoBERTa & Adjacent Sentence-Language & 64 & 1.04E-04 & 5 & 0.73442 & 0.73458 & 0.73485 & 0.73437\\
 & Without Language Tags & 64 & 4.80E-05 & 4 & 0.73288 & 0.73338 & 0.73281 & 0.73270\\
\midrule
 & Interleaved Word-Language & 64 & 6.25E-05 & 5 & 0.75474 & 0.75448 & 0.75468 & 0.75454\\
HingRoBERTa-Mixed & Adjacent Sentence-Language & 32 & 6.21E-05 & 5 & 0.73306 & 0.73297 & 0.73326 & 0.73295\\
 & Without Language Tags & 64 & 6.85E-05 & 2 & 0.71096 & 0.71114 & 0.71092 & 0.71087\\
\midrule
 & Interleaved Word-Language & 64 & 2.24E-05 & 5 & 0.68835 & 0.68973 & 0.68605 & 0.68583\\
Multilingual BERT & Adjacent Sentence-Language & 64 & 5.52E-05 & 7 & 0.72358 & 0.72325 & 0.72333 & 0.72328\\
 & Without Language Tags & 64 & 1.00E-04 & 4 & 0.67260 & 0.67382 & 0.67248 & 0.67193\\
\midrule
 & Interleaved Word-Language & 64 & 4.14E-05 & 5 & 0.69106 & 0.69101 & 0.68972 & 0.68985\\
AlBERT & Adjacent Sentence-Language & 64 & 2.84E-05 & 7 & 0.70461 & 0.70476 & 0.70500 & 0.70455\\
 & Without Language Tags & 64 & 5.79E-06 & 4 & 0.59863 & 0.60141 & 0.59885 & 0.59617\\
\midrule
 & Interleaved Word-Language & 64 & 9.20E-05 & 5 & 0.72629 & 0.72603 & 0.72623 & 0.72609\\
BERT & Adjacent Sentence-Language & 64 & 2.58E-05 & 7 & 0.71816 & 0.71812 & 0.71704 & 0.71724\\
 & Without Language Tags & 64 & 9.36E-05 & 4 & 0.61644 & 0.61648 & 0.61646 & 0.61643\\
\midrule
 & Interleaved Word-Language & 32 & 3.14E-05 & 7 & 0.69919 & 0.69892 & 0.69910 & 0.69897\\
RoBERTa & Adjacent Sentence-Language & 64 & 7.14E-05 & 5 & 0.69648 & 0.69610 & 0.69610 & 0.69610\\
 & Without Language Tags & 32 & 1.02E-05 & 5 & 0.66301 & 0.66369 & 0.66292 & 0.66259\\
\bottomrule
\end{tabular*}
\end{table}

\begin{table}[!ht]
\subsection{Emotions Dataset}
\caption{Emotions Results}\label{tab:EmotionsResults}
\begin{tabular*}{\tblwidth}{@{}C@{}C@{}C@{}C@{}C@{}C@{}C@{}C@{}C}
\toprule
Models & Approach & Batch Size & Learning Rate & Epochs & Accuracy & Precision & Recall & F1 Score\\ % Table header row
\midrule
 & Interleaved Word-Language & 64 & 1.73E-05 & 3 & 0.98268 & 0.98032 & 0.98130 & 0.98080\\
HingBERT & Adjacent Sentence-Language & 64 & 5.47E-05 & 2 & 0.98542 & 0.98296 & 0.98414 & 0.98353\\
 & Without Language Tags & 64 & 9.20E-05 & 1 & 0.82381 & 0.81890 & 0.81562 & 0.81437\\
\midrule
 & Interleaved Word-Language & 64 & 2.93E-05 & 3 & 0.98577 & 0.98370 & 0.98482 & 0.98425\\
HingRoBERTa & Adjacent Sentence-Language & 64 & 4.93E-05 & 2 & 0.98537 & 0.98292 & 0.98439 & 0.98362\\
 & Without Language Tags & 32 & 1.10E-05 & 4 & 0.98401 & 0.98177 & 0.98215 & 0.98195\\
\midrule
 & Interleaved Word-Language & 64 & 3.95E-05 & 5 & 0.98599 & 0.98427 & 0.98466 & 0.98447\\
HingRoBERTa-Mixed & Adjacent Sentence-Language & 64 & 3.48E-05 & 5 & 0.98515 & 0.98356 & 0.98359 & 0.98357\\
 & Without Language Tags & 64 & 4.84E-05 & 3 & 0.98480 & 0.98267 & 0.98306 & 0.98285\\
\midrule
 & Interleaved Word-Language & 32 & 3.97E-05 & 5 & 0.97771 & 0.97533 & 0.97630 & 0.97581\\
Multilingual BERT & Adjacent Sentence-Language & 32 & 1.52E-05 & 4 & 0.98432 & 0.98170 & 0.98300 & 0.98233\\
 & Without Language Tags & 32 & 7.25E-05 & 3 & 0.97960 & 0.97587 & 0.97841 & 0.97705\\
\midrule
 & Interleaved Word-Language & 64 & 4.08E-05 & 3 & 0.97506 & 0.97319 & 0.97329 & 0.97316\\
AlBERT & Adjacent Sentence-Language & 32 & 7.03E-05 & 7 & 0.97515 & 0.97195 & 0.97326 & 0.97255\\
 & Without Language Tags & 64 & 6.18E-05 & 4 & 0.95867 & 0.95502 & 0.95619 & 0.95540\\
\midrule
 & Interleaved Word-Language & 32 & 5.68E-05 & 3 & 0.98246 & 0.97976 & 0.98124 & 0.98047\\
BERT & Adjacent Sentence-Language & 32 & 6.26E-05 & 2 & 0.98445 & 0.98133 & 0.98332 & 0.98228\\
 & Without Language Tags & 32 & 5.81E-05 & 2 & 0.81271 & 0.80510 & 0.80702 & 0.80423\\
\midrule
 & Interleaved Word-Language & 64 & 6.73E-05 & 3 & 0.98211 & 0.97959 & 0.98084 & 0.98017\\
RoBERTa & Adjacent Sentence-Language & 64 & 6.62E-05 & 7 & 0.98374 & 0.98043 & 0.98268 & 0.98149\\
 & Without Language Tags & 32 & 4.07E-05 & 4 & 0.98264 & 0.97918 & 0.98091 & 0.98001\\
\bottomrule
\end{tabular*}
\end{table}

\twocolumn

% \bibliographystyle{abbrvnat}
% \bibliography{cas-refs}
\bibliographystyle{cas-model2-names} 
\bibliography{main}
% \printbibliography
% \twocolumn
% % Here goes the biography details.

% % Here goes the biography details.

\end{document}